\documentclass{article}

\PassOptionsToPackage{numbers, compress}{natbib}
\usepackage[preprint]{nips_2018}
\setcitestyle{number}

\usepackage{microtype}
\usepackage{graphicx}
\usepackage{subcaption}
\usepackage{booktabs} % for professional tables

\usepackage[utf8]{inputenc} % allow utf-8 input
\usepackage[T1]{fontenc}    % use 8-bit T1 fontss
\usepackage{amsfonts}       % blackboard math symbols
\usepackage{nicefrac}       % compact symbols for 1/2, etc.
\usepackage{times}
\usepackage{amsmath,amssymb}
\usepackage{amsfonts}
\usepackage{graphicx}
\usepackage{float}
\usepackage{color}
\usepackage[hyphenbreaks]{breakurl}
\usepackage[hyphens]{url}
\usepackage{tikz}
\usetikzlibrary{chains, positioning,decorations.markings}
\usepackage{comment}

\usepackage{hyperref}       % hyperlinks
\usepackage{xcolor}
\hypersetup{
    colorlinks,
    linkcolor={blue!50!black},
    citecolor={blue!50!black},
    urlcolor={blue!50!black}
}

\usepackage{xcite}

\usepackage{xr}
\makeatletter
\newcommand*{\addFileDependency}[1]{% argument=file name and extension
  \typeout{(#1)}% latexmk will find this if $recorder=0 (however, in that case, it will ignore #1 if it is a .aux or .pdf file etc and it exists! if it doesn't exist, it will appear in the list of dependents regardless)
  \@addtofilelist{#1}% if you want it to appear in \listfiles, not really necessary and latexmk doesn't use this
  \IfFileExists{#1}{}{\typeout{No file #1.}}% latexmk will find this message if #1 doesn't exist (yet)
}
\makeatother

\newcommand*{\myexternaldocument}[1]{%
    \externaldocument{#1}%
    \addFileDependency{#1.tex}%
    \addFileDependency{#1.aux}%
}
\myexternaldocument{supp}

\tikzset{dist/.style={path picture= {
    \begin{scope}[x=1pt,y=10pt]
      \draw[-] plot[domain=-5:5] (\x,{1/(1*sqrt(2*pi))*exp(-((\x-0)^2)/(2*1^2)) * 1.5 - 0.25});
    \end{scope}
    }
  }
}

\tikzset{dist2/.style={path picture= {
    \begin{scope}[x=0.5pt,y=10pt]
      \draw[-, color=orange] plot[domain=-8:8] (\x,{(1/(1*sqrt(2*pi))*exp(-((\x+3)^2)/(2*1^2)) + 1/(1*sqrt(2*pi))*exp(-((\x-3)^2)/(2*1^2))) * 1.5 - 0.25});
    \end{scope}
    }
  }
}

\tikzset{dist3/.style={path picture= {
    \begin{scope}[x=0.5pt,y=10pt]
      \draw[-, color=orange] plot[domain=-8:8] (\x,{(1/(1*sqrt(2*pi))*exp(-((\x+3)^2)/(2*1^2)) + 1/(1*sqrt(2*pi))*exp(-((\x-1)^2)/(2*1^2))) * 1.5 - 0.25});
    \end{scope}
    }
  }
}

\tikzset{dist4/.style={path picture= {
    \begin{scope}[x=0.5pt,y=10pt]
      \draw[-, color=orange] plot[domain=-8:8] (\x,{(1/(1*sqrt(2*pi))*exp(-((\x+3)^2)/(2*1^2)) + 1/(1*sqrt(2*pi*1^2))*exp(-((\x-3)^2)/(2*1^2)) + 1/(1*sqrt(2*pi*1^2))*exp(-((\x)^2)/(2*1^2))) * 1.5 - 0.25});
    \end{scope}
    }
  }
}

\tikzset{dist5/.style={path picture= {
    \begin{scope}[x=0.5pt,y=10pt]
      \draw[-, color=orange] plot[domain=-8:8] (\x,{1/(1*sqrt(2*pi))*exp(-((\x-0)^2)/(2*1^2)) * 1.5 - 0.25});
    \end{scope}
    }
  }
}

\tikzset{dist6/.style={path picture= {
    \begin{scope}[x=0.2pt,y=10pt]
      \draw[-, color=orange] plot[domain=0:50] (\x -20,{0.5*(\x/4)*(1-0.3*(1-\x/4))^(-1/0.3) - 0.1});
    \end{scope}
    }
  }
}

\tikzset{dist7/.style={path picture= {
    \begin{scope}[x=0.3pt,y=10pt]
      \draw[-, color=orange] plot[domain=0:20] (\x -10,{0.5*(\x/1)*(1-0.05*(1-\x/1))^(-1/0.05)});
    \end{scope}
    }
  }
}

\tikzset{dist8/.style={path picture= {
    \begin{scope}[x=1pt,y=10pt]
      \draw[-, color=orange] plot[domain=-6:6] (\x,{1/(1*sqrt(2*pi*3^2))*exp(-((\x-0)^2)/(2*3^2)) * 3 - 0.25});
    \end{scope}
    }
  }
}

\tikzset{dist9/.style={path picture= {
    \begin{scope}[x=0.1pt,y=20pt]
      \draw[-, color=orange] plot[domain=0:50] (\x - 25,{0.5*(\x/1)*(1-0.3*(1-\x/1))^(-1/0.3) - 0.15});
    \end{scope}
    }
  }
}

\definecolor{deepmagenta}{rgb}{0.8, 0.0, 0.8}

\definecolor{deepskyblue}{rgb}{0.0, 0.74901960784, 1.0}
\newcommand{\matt}[1]{{\textcolor{deepskyblue}{[#1]}}}
\definecolor{red}{rgb}{0.8, 0.0, 0.0}

\definecolor{green}{rgb}{0.0, 0.8, 0.0}

\definecolor{torqouis}{rgb}{0.0, 0.6, 0.6}
\newcommand{\nick}[1]{{\textcolor{torqouis}{[#1]}}}

\usepackage{hyperref}

\title{Implicit Weight Uncertainty in Neural Networks}

\usepackage{authblk}

\author[1]{Nick Pawlowski}
\author[2]{Andrew Brock}
\author[1,3]{Matthew C.H. Lee}
\author[1]{Martin Rajchl}
\author[1]{Ben Glocker}
\affil[1]{Biomedical Image Analysis Group, Imperial College London, UK}
\affil[2]{School of Engineering and Physical Sciences, Heriot-Watt University Edinburgh, UK}
\affil[3]{HeartFlow, California, CA 94063, USA}
\affil[ ]{\url{n.pawlowski16@imperial.ac.uk}}

\usepackage{todonotes}
\begin{document}

\maketitle
\begin{abstract}
% The abstract is a brief summary of the paper, which needs to be written extremely well. Try to address the following points in your abstract, with a single sentence per point. This will naturally keep the abstract compact:
% 1. Describe the task/problem the paper is going to address (high level)
% 2. Why is this an interesting/important problem?
% 3. How does one usually solve this?
% 4. How (and why) do we do it in this paper (key idea)? Highlight the novelty here.
% 5. Interpretation of the results (impact and importance)
%\nick{When editing sections please indicate with \textbackslash name\{bla\} that you are working on this section}
Modern neural networks tend to be overconfident on unseen, noisy or incorrectly labelled data and do not produce meaningful uncertainty measures. Bayesian deep learning aims to address this shortcoming with variational approximations (such as Bayes by Backprop \cite{blundell2015weight} or Multiplicative Normalising Flows \cite{louizos2017multiplicative}). However, current approaches have limitations regarding flexibility and scalability. We introduce Bayes by Hypernet (\emph{BbH}), a new method of variational approximation that interprets hypernetworks \cite{ha2016hypernetworks} as implicit distributions. It naturally uses neural networks to model arbitrarily complex distributions and scales to modern deep learning architectures. In our experiments, we demonstrate that our method achieves competitive accuracies and predictive uncertainties on MNIST and a CIFAR5 task, while being the most robust against adversarial attacks. %these aspects with a ResNet-32 architecture and report competitive accuracy with good predictive uncertainty.
% We interpret \textit{HyperNetworks} \cite{ha2016hypernetworks} within the framework of variational inference within implicit distributions\cite{tran2017deep,huszar2017variational,mescheder2017adversarial}. Our method, Bayes by Hypernet, is able to model a richer variational distribution than previous methods. Experiments show that it achieves comparable predictive performance and uncertainty on a MNIST and CIFAR classification task. We show that our method is better scalable to modern neural network architectures like ResNets.
\end{abstract}
%\nick{keep reviews in mind and maybe look at \url{https://media.nips.cc/nipsbooks/nipspapers/paper_files/nips30/reviews/3205.html}}

\section{Introduction}
Neural networks achieve state of the art results on a wide variety of tasks \cite{lecun2015deep}, with applications spanning image recognition \cite{hu2017}, machine translation \cite{lample2018unsupervised} and reinforcement learning \cite{silver2017mastering}. Such success is often mitigated by the need for vast troves of data, and a tendency towards poorly calibrated and overconfident predictions \cite{guo2017calibration}. However, real-world decision making processes that wish to leverage neural networks (\emph{e.g.} medical applications, self-driving cars, \emph{etc.}) are frequently faced with a dearth of data and the need for reliable uncertainty estimates, as overconfidence in the wrong situation could prove dangerous \cite{amodei2016concrete}.

To address the issue of overconfident predictions, recent works have proposed approaches like calibration methods \cite{guo2017calibration}, frequentist interpretations of ensembles \cite{lakshminarayanan2017simple,osband2016deep}, and approximate Bayesian inference \cite{welling2011bayesian,mackay1992practical}. Of those approaches, the work on Bayesian deep learning (BDL) offers a particularly principled approach to enable uncertainty estimates within the existing deep learning framework as it aims to marginalise the model parameters. 
\begin{figure*}[h]
\centering
	\includegraphics[width=\textwidth]{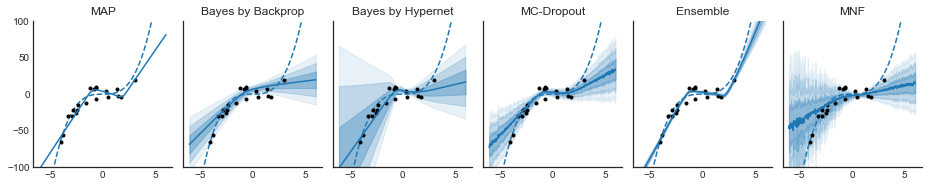}
	\caption{Toy example inspired by \cite{hernandez2015probabilistic}: Real function (dashed line) with sampled data points (black dots). The proposed \emph{BbH} exhibits the best trade-off between predictive uncertainty and regression fit. MC-Dropout \cite{gal2015bayesian}, deep ensembles \cite{lakshminarayanan2017simple} and the MAP produce a good fit but underestimate the predictive uncertainty. Multiplicative Normalizing Flows (MNF) \cite{louizos2017multiplicative} and Bayes by Backprop (BbB) \cite{blundell2015weight} achieve a slightly worse fit and predictive uncertainty than \emph{BbH}.}
   \label{fig:toy_example}
\end{figure*}

The current research in BDL is primarily divided into variational inference methods \cite{blundell2015weight,gal2015bayesian,louizos2017multiplicative} and Monte Carlo methods \cite{welling2011bayesian,chen2014stochastic,lu2017relativistic}. A simple toy example (Fig.~\ref{fig:toy_example}) illustrates the issue of low predictive uncertainty in unseen regions for regular deep learning methods as well as the more reliable uncertainty of Bayesian approximations.

Applying approximate Bayesian inference with neural networks was first studied by MacKay \cite{mackay1992practical} and Neal \cite{neal1995phd}. Both remain relevant today with the Laplace approximation \cite{mackay1992practical} being one of the easiest to use Bayesian approximations to date and Hamiltonion Monte Carlo \cite{neal1995phd} being one of the most widely employed Monte-Carlo methods. However, both methods scale poorly to current neural network architectures due to the computational burden induced by the high dimensionality of the weight space. 

\textbf{Related work:} More recent methods improve the scalability of approximate inference methods and the complexity of the variational approximation, introducing sampling-based methods like SGLD \cite{welling2011bayesian} or other minibatch-based sampling methods \cite{ma2015complete}. Graves \cite{graves2011practical} proposed a simple, but biased method to perform variational inference with a fully factorized posterior distribution. This work was extended by \cite{blundell2015weight} using the reparametrisation trick from \cite{kingma2013auto} and scale-mixture priors. An Expectation Propagation \cite{minka2001expectation} based approach using a fully factorized posterior approximation was proposed by \cite{hernandez2015probabilistic}. Dropout-based \cite{srivastava2014dropout} approximate inference methods have been proposed by \cite{kingma2015variational} using Gaussian Dropout and by \cite{gal2015bayesian} using Bernoulli Dropout. Further, \cite{louizos2016structured} introduced structured posterior approximations that use matrix Gaussians rather than fully-factorized Gaussians as in \cite{blundell2015weight}.

Several studies \cite{louizos2017multiplicative,krueger2017bayesian} proposed to employ normalising flows to further increase the flexibility of the variational approximation\footnote{Krueger et al. \cite{krueger2017bayesian} use the term \emph{hypernetwork} to refer to normalising flows rather than the more general concept of weight generating networks from \cite{ha2016hypernetworks}.}. However, both approaches only employ the high-fidelity approximations as multiplicative factors on otherwise factorised Gaussians \cite{louizos2017multiplicative} or single delta peaks \cite{krueger2017bayesian}. Only recently, implicit models have been studied under the framework of variational inference \cite{mescheder2017adversarial,huszar2017variational,tran2017deep}, but only \cite{shi2018kernel} have used implicit distributions to model weight uncertainty. However, the proposed method parametrises weight matrices as outer product of two vectors and is therefore limited in it expressibility. %\martin{Need to be more explicit about the limitation of \cite{shi2018kernel}}
%The closest approach to \textit{BbH} was published by , where Multiplicative Normalizing Flows (MNF) are employed to increase the flexibility of the variational approximation. This allows for the method to better fit the real posterior, thus increasing the performance. However, MNF requires the use of invertible transformations to calculate the log density ratio (see \autoref{eq:ratio}) of the KL-divergence and need to compute the Jacobians of those transformations. Furthermore, it requires a lower bound to the variational lower bound as their parametrisation does not have a closed form density. In comparison, our work also does not have a closed form density, but can exploit the full potential of neural networks to approximate the posterior distribution. \matt{Need to change the following: Rather than bounding the lower bound we rely on adversarial training to perform variational inference. Therefore, our method suffers from known instabilities in GAN training \cite{arjovsky2017towards}, whereas MNF only require a single step for training which makes it more stable. A special case of MNF was published by \cite{krueger2017bayesian}.}

In contrast to Bayesian inference methods, frequentist approaches have recently been proposed. A bootstrap-based approach was proposed by \cite{osband2016deep}, whereas \cite{lakshminarayanan2017simple} use ensembles of deep networks to calculate predictive uncertainties based on the sample difference due to different initialisation and noise in the stochastic gradients. Even though deep ensembles are straight forward to train their main disadvantages are that the computational cost scales linearly with the amount of networks in the ensemble and that their uncertainty estimation solely relies on noise during training rather than being principled like Bayesian methods.%\todo{please someone read this again}

\textbf{Contributions: } We propose to combine prior work on implicit variational inference \cite{mescheder2017adversarial,huszar2017variational,tran2017deep,shi2018kernel} with the concept of hypernetworks \cite{ha2016hypernetworks}. This builds Bayes by Hypernet as we reinterpret hypernetworks \cite{ha2016hypernetworks} as implicit distributions similar to generators in generative adversarial networks \cite{goodfellow2014generative} and use them to approximate the posterior distribution of the weights of a neural network. Hypernetworks are able to model a wide range of distributions and can therefore provide rich variational approximations. Furthermore, the hypernetwork is inherently learning complex correlations between the weights as it generates samples of multiple weights at the same time. \emph{Bayes by Hypernet} (\emph{BbH}) avoids hand-crafted strategies of building variational approximations and instead exploits the inherent capabilities of learned approximations to model rich, varied distributions. We show that compared to other Bayesian methods \emph{BbH} achieves competitive performance on small networks and its computational cost scales better to modern deep architectures. \emph{BbH} demonstrates comparable predictive accuracy without compromising predictive uncertainty, while being the least vulnerable against adversarial attacks.

\section{Bayes by Hypernet}
\noindent\textbf{Variational Bayesian Neural Networks: }
Given a dataset $\mathcal{D}$ with data points ${(x_1, y_1), \dots, (x_n, y_n)}$ variational inference for Bayesian neural networks aims to approximate the posterior distribution $p\left(\mathbf{w}\mid \mathcal{D}\right)$ of the weights $\mathbf{w}$ of a neural network. Given this distribution we can estimate the posterior prediction $\hat{y}$ of a new data point $\hat{x}$ as $p\left(\hat{y}\mid \hat{x}, \mathcal{D}\right) = \mathbb{E}_{w\sim p\left(\mathbf{w}\mid \mathcal{D}\right)}\left[p\left(\hat{y}\mid \hat{x}, \mathbf{w} \right) \right]$. Because exact Bayesian inference is usually intractable in neural networks we find a variational approximation $q\left(\mathbf{w}\mid\mathbf{\theta}\right)$ with parameters $\theta$ that minimises the evidence lower bound (ELBO):%Kullback-Leibler (KL) divergence:
\begin{align}
\theta^{*} &= \arg\min_{\theta} KL\left(q\left(\mathbf{w}\mid\mathbf{\theta}\right) \| \, p\left(\mathbf{w}\mid \mathcal{D}\right)\right)&& \nonumber\\
&= \arg\min_{\theta}  KL\left(q\left(\mathbf{w}\mid\mathbf{\theta}\right)\| \, p\left(\mathbf{w}\right)\right) - \mathbb{E}_{w\sim q\left(\mathbf{w}\mid\mathbf{\theta}\right)}\left[\log p\left(\mathcal{D}\mid \mathbf{w}\right)\right]&& \nonumber\\
&= \arg\min_{\theta} \mathbb{E}_{w\sim q\left(\mathbf{w}\mid\mathbf{\theta}\right)}\left[\log \frac{q\left(\mathbf{w}\mid\mathbf{\theta}\right)}{p\left(\mathbf{w}\right)}  - \log p\left(\mathcal{D}\mid \mathbf{w}\right) \vphantom{\frac{q\left(\mathbf{w}\mid\mathbf{\theta}\right)}{p\left(\mathbf{w}\right)}} \right] &&\label{eq:ratio}
\end{align} 
Recent works improved upon the Laplace approximation \cite{mackay1992practical} by using the reparametrisation trick \cite{kingma2013auto} or stochastic backpropagation \cite{rezende2014stochastic}. One of the first works to combine the reparametrisation trick with variational inference for Bayesian neural networks used fully factorised Gaussians \cite{blundell2015weight} to model the approximative distribution. This allows for straightforward optimisation but incurs strong limitations by allowing only unimodal distributions in the high dimensional weight space of neural networks.

\noindent\textbf{More complex variational approximations: }
Various works have since proposed different extensions to allow for richer approximations such as mixture of delta peaks \cite{gal2015bayesian} or Matrix Gaussians \cite{louizos2016structured}. Nevertheless, those approximations are far from optimal as the true posterior will  likely be more complex than delta peaks or correlated Gaussians \cite{louizos2017multiplicative}. Recently, normalising flows have been proposed to allow for more complex approximative distributions \cite{louizos2017multiplicative,krueger2017bayesian}. Normalising flows use bijective functions with learnable parameters and simple Jacobians to transform samples from simple densities into more complex distributions. By stacking multiple of those like the layers of neural networks it is possible to resemble highly complex distributions \cite{rezende2015variational}. However, both Multiplicative Normalising Flows (MNF) and Bayesian Hypernetworks \cite{krueger2017bayesian} only use normalising flows as multiplicative factors of variation and model the majority of the weights with factorised Gaussians \cite{louizos2017multiplicative} or regular point estimates \cite{krueger2017bayesian}. This parametrisation limits the relations between weights that are able to be modelled and the parametrisation of \cite{louizos2017multiplicative} requires an auxiliary inference network.

\noindent\textbf{Hypernetworks as Implicit Distributions: }
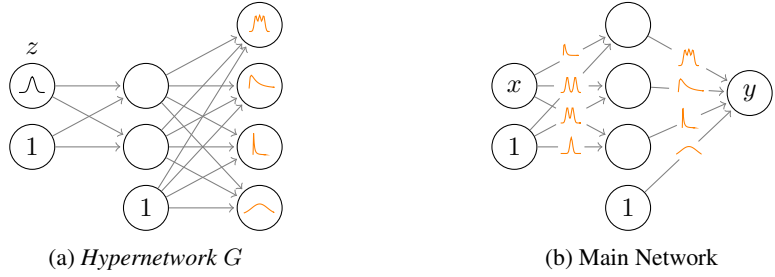
\begin{figure*}
\centering
\begin{subfigure}[b]{0.3\textwidth}
\centering
\begin{tikzpicture}[shorten >=1pt,->, draw=black!50, 
        node distance = 2mm and 9mm,
          start chain = going below,
every pin edge/.style = {<-,shorten <=1pt},
        neuron/.style = {circle, fill=#1, draw=black,
                         minimum size=17pt, inner sep=0pt,
                         on chain},
         annot/.style = {text width=4em, align=center}
         decoration={
    markings,
    mark=between positions 0.25 and 0.75 step 0.125 with {\node [yshift=0.3cm] {$y_\pgfkeysvalueof{/pgf/decoration/mark info/sequence number}$};},
    mark=between positions 0.125 and 0.875 step 0.125 with {\fill (0pt,0pt) circle (2pt);},
  }
                        ]
% Draw the input layer nodes

    \node[neuron=white,dist, label=$z$] (I-1)    {};
    \node[neuron=white,
          below=of I-1] (I-2)     {$1$};
% Draw the hidden layer nodes
    
    \node[neuron=white,
          right=of I-1] (H-1)     {};
    \node[neuron=white,
          below=of H-1] (H-2)     {};
    \node[neuron=white,
          below=of H-2] (H-4)     {$1$};
% Draw the output layer node
    \node[neuron=white,
          right=of H-2, dist9]  (O-1)   {};
    \node[neuron=white,
          above=of O-1, dist6]  (O-4)   {};
    \node[neuron=white,
          below=of O-1, dist8]  (O-2)   {};
    \node[neuron=white, %label=$w_2$,
          above=of O-4, dist4]  (O-3)   {};
% Connect input nodes with hidden nodes and 
%  hiden nodes with output nodes with the output layer
\path (I-1) edge (H-1) (H-1) edge (O-1);
\path (I-1) edge (H-2) (H-2) edge (O-1);

\path (H-4) edge (O-1) ;
\path (I-2) edge (H-1) ;
\path (I-2) edge (H-2) ;

\path (H-1) edge (O-2) ;
\path (H-2) edge (O-2) ;
\path (H-4) edge (O-2) ;

\path (H-1) edge (O-3) ;
\path (H-2) edge (O-3) ;
\path (H-4) edge (O-3) ;

\path (H-1) edge (O-4) ;
\path (H-2) edge (O-4) ;
\path (H-4) edge (O-4) ;

    \end{tikzpicture}
    \caption{\textit{Hypernetwork} $G$}
    \label{fig:bbh_hnet}
\end{subfigure}
\hspace{0.15\textwidth}
\begin{subfigure}[b]{0.3\textwidth}
\centering
\begin{tikzpicture}[shorten >=1pt,->, draw=black!50, 
        node distance = 2mm and 9mm,
          start chain = going below,
every pin edge/.style = {<-,shorten <=1pt},
        neuron/.style = {circle, fill=#1, draw=black,
                         minimum size=17pt, inner sep=0pt,
                         on chain},
         annot/.style = {text width=4em, align=center},
  		 edge_node1/.style = {midway, dist2,minimum size=13pt,fill=white, circle},
         edge_node2/.style = {midway, dist3,minimum size=13pt,fill=white, circle},
         edge_node3/.style = {midway, dist6,minimum size=15pt,fill=white, circle},
         edge_node4/.style = {midway, dist5,minimum size=10pt,fill=white, circle},
         edge_node5/.style = {midway, dist4,minimum size=15pt,fill=white, circle},
         edge_node6/.style = {midway, dist7,minimum size=10pt,fill=white, circle},
         edge_node7/.style = {midway, dist8,minimum size=10pt,fill=white, circle},
         edge_node8/.style = {midway, dist9,minimum size=10pt,fill=white, circle}
                        ]
         
% Draw the input layer nodes

    \node[neuron=white] (I-1)    {$x$};
    \node[neuron=white] (I-2)    {$1$};
% Draw the hidden layer nodes
    
    \node[neuron=white,
          right=of I-1] (H-2)     {};
    \node[neuron=white,
          above=of H-2] (H-1)     {};
    \node[neuron=white,
          below=of H-2] (H-3)     {};
    \node[neuron=white,
          below=of H-3] (H-4)     {$1$};
% Draw the output layer node
    \node[neuron=white,
          above right=5mm and 14mm of H-3.center]  (O-1)   {$y$};
% Connect input nodes with hidden nodes and 
%  hiden nodes with output nodes with the output layer
\path (I-1) edge node [edge_node6] {} (H-1);
\path (H-1) edge node [edge_node5] {} (O-1);
\path (I-1) edge node [edge_node1] {} (H-2);
\path (H-2) edge node [edge_node3] {} (O-1);
\path (I-1) edge node [edge_node2] {} (H-3);
\path (H-3) edge node [edge_node8] {} (O-1);

\path (I-2) edge node [edge_node1] {} (H-1) ;
\path (I-2) edge node [edge_node2] {} (H-2) ;
\path (I-2) edge node [edge_node4] {} (H-3) ;

\path (H-4) edge node [edge_node7] {} (O-1) ;

    \end{tikzpicture}
    \caption{Main Network}
    \label{fig:bbh_net}
\end{subfigure}
\caption{Illustration of the components of Bayes by Hypernet: The \textit{hypernetwork} $G$ takes a sample $z \sim p(z)$ and converts it into a sample of the weights $w$ of the main network. The \textit{hypernetwork} in Fig.~\ref{fig:bbh_hnet} generates samples of the weights of the second layer of the main network. The main network takes a data sample $x$ and generates an output $y$ using the weight samples generated by the \textit{hypernetworks}.
%The discriminator sees weight samples $w$ from the prior, $w \sim p(w)$, as well as generated sample from the \textit{hypernetworks}, $w \sim G(z\mid \theta)$. The discriminator is optimised to distinguish between the two types of samples. The generator is optimised to fool the discriminator and provide high likelihood of the training data using the main network.
}
\label{fig:bbh_tikz}

\end{figure*}
Implicit distributions are distributions that may have intractable probability densities but allow for easy sampling. They enable simple calculation of approximate expectations and their corresponding gradients \cite{huszar2017variational}. Probably the most well-known group of implicit distributions are generative adversarial networks \cite{goodfellow2014generative} that can transform a sample from a simple noise distribution into high-fidelity images.

Using an implicit distribution to model the weights of a neural network requires a generator that is able to capture inherent complexity of neural networks weights. Hypernetworks \cite{ha2016hypernetworks} are shown to be able to generate weights of networks like ResNets or RNNs while still achieving competitive state-of-the-art performances. Let $G$ be a hypernetwork with parameters $\theta$. Further, let $z$ be an input vector to the hypernetwork $G$ that contains information about the weight $w$ to generate. Then weights $w$ of the main network are generated as $w = G(z\mid \theta)$.

When $z$ is a sample from a simple auxiliary random variable the hypernetwork resembles a generator within the GAN framework. Rather than generating high-fidelity image samples, our generator predicts samples of the weight distribution of the main network. An illustration of the combination of hypernetwork and main network is shown in Figure \ref{fig:bbh_tikz}. The graphs shown resemble distributions the auxiliary variable $z$ or the weight samples $w$ could be drawn from.%\todo{reread previous paragraph}

In the original work \cite{ha2016hypernetworks}, hypernetworks were introduced as means of weight sharing and therefore network compression. Here, we do not focus on compression, but use arbitrary neural networks as hypernetworks. This is different from the terminology presented in \cite{krueger2017bayesian} where stacked normalising flows are called hypernetworks.
% present different parametrisations in experimentation section

\noindent\textbf{Estimating the ELBO: }
Because implicit distributions do not have tractable probability densities, the prior matching term $KL(q(w\mid\theta)|| p(w))$ of the ELBO becomes intractable. Previous works \cite{mescheder2017adversarial,huszar2017variational,tran2017deep} describe how to perform variational inference with implicit distributions. The proposed approaches closely follow the structure of adversarial training, where a generator $w = G(z\mid \mathbf{\theta})$ models the variational distribution $q\left(\mathbf{w}\mid\mathbf{\theta}\right)$ and a discriminator $D$ estimates the log density ratio from Eq. \ref{eq:ratio}. Here, $z$ is an auxiliary noise variable $z \sim p(z)$ which may also contain additional conditioning information.

\begin{comment}
Specifically, we follow the notion of prior-contrastive adversarial variational inference as formulated by \cite{huszar2017variational} and estimate the density ratio in \autoref{eq:ratio} using logistic regression. This enables a two-step update procedure with 
\begin{align}
\mathcal{L}\left(D\mid G\right) &= \mathbb{E}_{w\sim G(z\mid \mathbf{\theta})} \log D(\mathbf{w}) + \mathbb{E}_{w\sim p(\mathbf{w})} \log \left(1 - D(\mathbf{w})\right) \label{eq:discloss}\\
\mathcal{L}\left(G\mid D\right) &= \mathbb{E}_{z} \log \frac{D\left(G(z\mid \mathbf{\theta})\right)}{1 - D\left(G(z\mid \mathbf{\theta})\right)} - \mathbb{E}_{z} \log p\left(\mathcal{D}\mid G(z\mid \mathbf{\theta})\right)\label{eq:genloss}
\end{align}
where \autoref{eq:discloss} and \autoref{eq:genloss} are being used to update the discriminator and the generator, respectively\footnote{See \textit{Variational Inference and Density Ratios} section on \url{https://tinyurl.com/zpeaboh}}. In theory, the discriminator only gives exact gradients when it has converged to an optimal solution, but GAN training shows us that non-converged discriminators can still provide useful gradients.
\end{comment}

However, it is not straightforward to employ adversarial training in high-dimensional spaces such as neural network weights which can accrue to 100 thousands or millions of parameters\footnote{The LeNet that we use in a later experimental section already has more than 400,000 weights. In comparison, most image datasets used in deep learning do not have more than 196,608 ($256 \times 256 \times 3$) dimensions.}. This number of input dimensions raises computational issues as it spans huge weight matrices when dense layers are employed. We therefore propose to treat all weights independently and estimate the density ratio by a single discriminator. We compare this approximation to the analytical form of Bayes by Backprop (BbB) \cite{blundell2015weight} and find that the single discriminator is not capable of estimating the density ratios correctly%\footnote{For more information on this comparison see the supplementary material.}
. Instead we find that estimating the density ratio using a kernel method \cite{perez2008kullback} yields results that are close those of the analytical method. Specifically, we use the formulation from \cite{pmlr-v84-jiang18a}, approximating the KL divergence $KL(q(w\mid\theta)||p(w))$ as
\begin{align}
KL(q(w\mid\theta)||p(w)) = \frac{d}{n}\sum\limits_{i=1}^{n}\log\frac{\min_j||w_q^i - w_p^j||}{\min_{j\neq i}||w_q^i - w_q^j||} + \log\frac{m}{n - 1} , \label{eq:kernel_kl}
\end{align}
where $d$ is the dimensionality of the samples $w$, $n$ is the number of approximate samples, $m$ is the number of prior samples, and $w_q$ and $w_p$ are samples from the approximative posterior and prior respectively. This resembles a ratio of distance between the nearest neighbours. On the same BbB task we find that this approximation produces results close to the analytical algorithm. Further, due to the nature of the independent Gaussian prior we propose to treat each weight independently as samples of $d=1$ and find on BbB that this achieves better performance than using this estimation on the full weight matrix\footnote{More information on adversarial training, kernel estimation and the independent treatment of weights can be found in Sec. \ref{sec:supp_bbb} in the supplementary material.}. We therefore use the kernel estimation treating each weight independently throughout the rest of this paper.

\section{Experiments}
We aim to assess the predictive accuracy of a method, and also its ability to estimate the predictive uncertainty. We closely follow established benchmarks \cite{louizos2017multiplicative} by comparing the performance of \emph{BbH} on the MNIST digit classification task and an adaptation on CIFAR10 classification. Additionally to accuracy, we test the entropy of the softmax outputs as a measure of predictive uncertainty and the method's robustness against adversarial examples. Methods are compared in their predictive uncertainty on test set (in-dataset examples) and on similar, yet unseen data (out-of-dataset examples). An optimal method would predict low uncertainty and correct predictions for the in-dataset examples and high uncertainty for out-of-dataset examples. The high uncertainty predictions on unseen data can be important in real-life decision processes as they can be used to trigger a request for human support. Similarly, by testing the robustness against adversarial attacks, we expect to see the degradation of accuracy and a simultaneous increase in predictive uncertainty. In contrast to \cite{louizos2017multiplicative, lakshminarayanan2017simple}, we do not rely on (cumulative) density plots of the entropy, but rather calculate the area under the curve (AUC) of the cumulative density plots to provide a quantitative measure. Here, we normalise via the maximum entropy of the classification problem.

We employ 3-layer fully-connected networks with $[64, 256, 512]$ units as hypernetworks for all experiments, as we did not find a general improvement by adding more layers or units. Further, we employ a standard normal prior for all methods (excluding ensembles) and treat all weights as independent. The experiments are implemented in Tensorflow \cite{abadi2016tensorflow} and optimization is performed with Adam \cite{kingma2014adam} with learning rate $\eta = 0.0001$ for \emph{BbH} and $\eta=0.001$ for the rest. We compare our method to MC-Dropout \cite{gal2015bayesian} (dropout rate $\pi = 0.5$), Bayes by Backprop (BbB) \cite{blundell2015weight}, deep ensembles \cite{lakshminarayanan2017simple}, multiplicative normalizing flows (MNF) \cite{louizos2017multiplicative}, and maximum a posteriori (MAP) training. We train the deep ensembles without predictive uncertainty as we found it to sometimes result in numerically unstable training. MNF and \emph{BbH} anneal the KL term during training. All methods use 100 posterior samples to estimate the predictive distribution and we use 5 samples to estimate the KL from Eq.~\ref{eq:kernel_kl}.
%All reported run times are for runs on nVidia Titan Xps using CUDA 8.
The source code to reproduce all experiments is available on \url{https://github.com/pawni/BayesByHypernet/}. % TODO
%\nick{explain the experiments better and why they are reasonable, change this:}
% We calculate the predictive entropy of the softmax output on notMNIST\footnote{Can be found at \url{http://yaroslavvb.blogspot.co.uk/2011/09/notmnist- dataset.html} We use the version from \url{https://github.com/davidflanagan/notMNIST-to-MNIST}} similar to \cite{louizos2017multiplicative, lakshminarayanan2017simple} to estimate the quality of predictive uncertainty obtained by all methods. Because the network is trained on MNIST (digits) rather than notMNIST (letters) the optimal prediction in this case would be a uniform output over all possible digits. This is equivalent to the maximum entropy solution which corresponds to an entropy of approximately $2.3$. We agree with \cite{louizos2017multiplicative} that density plots of entropy values as in \cite{lakshminarayanan2017simple} are hard to read. However, we found that the cumulative density of entropy values can be misleading because it is tightly related to the entropy on seen data. Therefore, we plot both the cumulative density of the entropy on MNIST and notMNIST in Figure \ref{fig:notmnist_cdf}. The plots show how many predictions have a smaller entropy than a given value $x$, so that a lower curve relates to a higher uncertainty. Additionally we quantify those plots by calculating the area under the curve of the cumulative density in $\left[0, 2.3\right]$ and normalise it by the maximum possible entropy to give a more objective metric. In general a well performing method would yield a low area under the curve for the outlier set and a high value for the in-dataset-examples.

\subsection{MNIST Digit Classification}
We reproduce the setup from \cite{louizos2017multiplicative}, employing a LeNet for MNIST classification and notMNIST as outlier dataset. Additionally from a comparison with other methods, we use MNIST to test newly introduced hyperparameters of the proposed method: The architecture of the hypernetwork, the composition of the auxiliary variable $z$ as input to the hypernetwork, and whether or not to use the same sample $z \sim p(z)$ across different weights that are generated at the same time.

\textbf{Architectures:} We compare three different parametrisations of the weights. First, we use a single hypernetwork $G_1$ that takes as input a vector $[z, c]$ where $z$ is the auxiliary noise and $c$ is a one-hot vector with $C$ entries encoding which part of the weights to generate. We can then generate all weights $W$ of the  main network by concatenating all generated weight slices given the specific conditioning $c$, $W = \{w_i = G([z, c])\} \mid z~\sim p(z) \forall c \in C $. Second, we employ a different hypernetwork for each of the $L$ layers. There the set of weights is given by $W = \{w_l = G_l(z)\}\mid z \sim p(z)~\forall l \in \{1\dots L\} $. Third, we combine those approaches and implement a single hypernetwork for each layer, only predicting slices of the relevant weight. Here, we chose to slice the weight of a layer $l$ along its $C_l$ output dimensions and the set of weights is $W = \{w_{i, l} = G_l([z, c])\} \mid z \sim p(z)~\forall c \in C_l~\forall l \in \{1\dots L\}$. We show the results in Table \ref{tab:bbh_arch}. We find that using a single hypernetwork per layer without slicing gives the best results. We argue, that this architecture has the biggest capacity to model different distributions and is therefore able to generate better weights for the task at hand. However, the slice-wise approaches allow for a faster generation as multiple weight slices can be batched. All further experiments use the layer-wise architecture.
%\matt{this is a pretty important section I would give it more space and break it up a bit more. Maybe open with the fact that weights of a neural network can be considered as a block of weights (1 hyper net to rule them all) or grouped by layer and then explain the different architectures w.r.t how we would group the weights semantically}

\begin{table}[tb]
\centering
\caption{Comparison of different hypernetwork architectures for \emph{BbH}: The layer-wise generation of weights achieves the best performance and a single hypernetwork results in the fastest runtime.\\}
\label{tab:bbh_arch}
\begin{small}
	\begin{tabular}{l r r r r}
	\toprule
 &  Error [$\%$]& MNIST AUC & Outlier AUC & Runtime [s]\\
 \midrule
 Single $G$ & $1.1$ & $0.89$ & $0.48$ & $2961$ \\
 Sliced layer-wise $G_l$ & $1.7$ & $0.93$ & $0.55$ & $7119$\\
 Layer-wise $G_l$ & $0.73$ & $0.98$ & $0.54$ & $11361$ \\
  \bottomrule
\end{tabular}
\end{small}
\end{table}

\textbf{Auxiliary Noise:} The hypernetworks take an auxiliary noise variable $z$ as an input and transform it into a sample of the weight distribution. In all experiments, we draw samples from unit-variate Gaussians as $z$. However, the dimensionality $d$ of $z$ can influence the capacities of the hypernetwork. Further, hypernetworks can be coupled by drawing the same sample $z$ for each weight or can be decoupled by drawing a different sample for each weight. The former enables the hypernetworks to learn more complicated relations across different parts of the weights, whereas the latter leads to higher variability across the generated weights. Table \ref{tab:bbh_noise} shows the results of different noise configurations. We find that a higher degree of noise (independent noise, higher dimensionality) increases the predictive uncertainty, but the relationship between these parameters and accuracy stays unclear.

\begin{table}[tb]
\centering
\caption{Comparison of different auxiliary noise configurations: A higher degree of noise increases the predictive uncertainty (lower outlier AUC) but do not demonstrate a trend in the corresponding accuracy.\\}
\label{tab:bbh_noise}
\begin{small}
	\begin{tabular}{l r r r r}
	\toprule
 &  Error [$\%$]& MNIST AUC & Outlier AUC & Runtime [s]\\
 \midrule
 Shared noise, $d=1$ & $0.73$ & $0.98$ & $0.54$ & $11361$ \\
 Independent noise, $d=1$ & $0.56$ & $0.97$ & $0.48$ & $11504$\\
 Shared noise, $d=8$ & $0.56$ & $0.98$ & $0.49$ & $11314$ \\
 Independent noise, $d=8$ & $0.64$ & $0.97$ & $0.44$ & $11268$\\
  \bottomrule
\end{tabular}
\end{small}
\end{table}

We further compare the performance of \emph{BbH} with independent noise with several established variational Bayesian deep learning techniques and frequentist approaches. The results are shown in Table \ref{tab:mnist}. All methods achieve comparable accuracy, with \emph{BbH} only being outperformed by deep ensembles and MC-Dropout. However, \emph{BbH} exhibits a higher predictive uncertainty, only outperformed by BbB on this metric. The runtime of \emph{BbH} and MNF are significantly increased over other approaches, because of a relative high overhead to generate the weights compared to the actual network architecture.

\begin{table}[tb]
\centering
\caption{Compared methods on the MNIST classification task: The Error [\%] is the methods classification error on the MNIST test set. The MNIST AUC and Outlier AUC refers to the AUC under the CDF of the predictive entropy on the corresponding data set. Compared methods: maximum a posteriori training (MAP), deep ensembles \cite{lakshminarayanan2017simple} (Ensembles), Bayes by Backprop \cite{blundell2015weight} (BbB), MC-Dropout \cite{gal2015bayesian} (Dropout), Multiplicative Normalizing Flows \cite{louizos2017multiplicative} (MNF) and Bayes by Hypernet (\emph{BbH}).\\}
\label{tab:mnist}
\begin{small}
	\begin{tabular}{l r r r r}
	\toprule
 &  Error [$\%$]& MNIST AUC & Outlier AUC & Runtime [s]\\
 \midrule
 MAP & $0.80$ & $ 0.99$   & $ 0.71$ & $710$ \\
 Ensemble & $0.49$ & $ 0.99$ & $ 0.65$& $2721$ \\
 BbB & $0.72$ & $ 0.97$ & $ 0.41$ & $2892$ \\
 Dropout & $ 0.47$ & $ 0.99$ & $ 0.58$ & $1224$ \\
 MNF & $ 0.63$ & $ 0.99$ & $ 0.58$ & $21811$ \\
 BbH (ours) & $ 0.56$ & $ 0.97$ & $ 0.48$ & $11504$ \\
  \bottomrule
\end{tabular}
\end{small}
\end{table}

\begin{figure}[tb]
\centering
	\includegraphics[width=\linewidth]{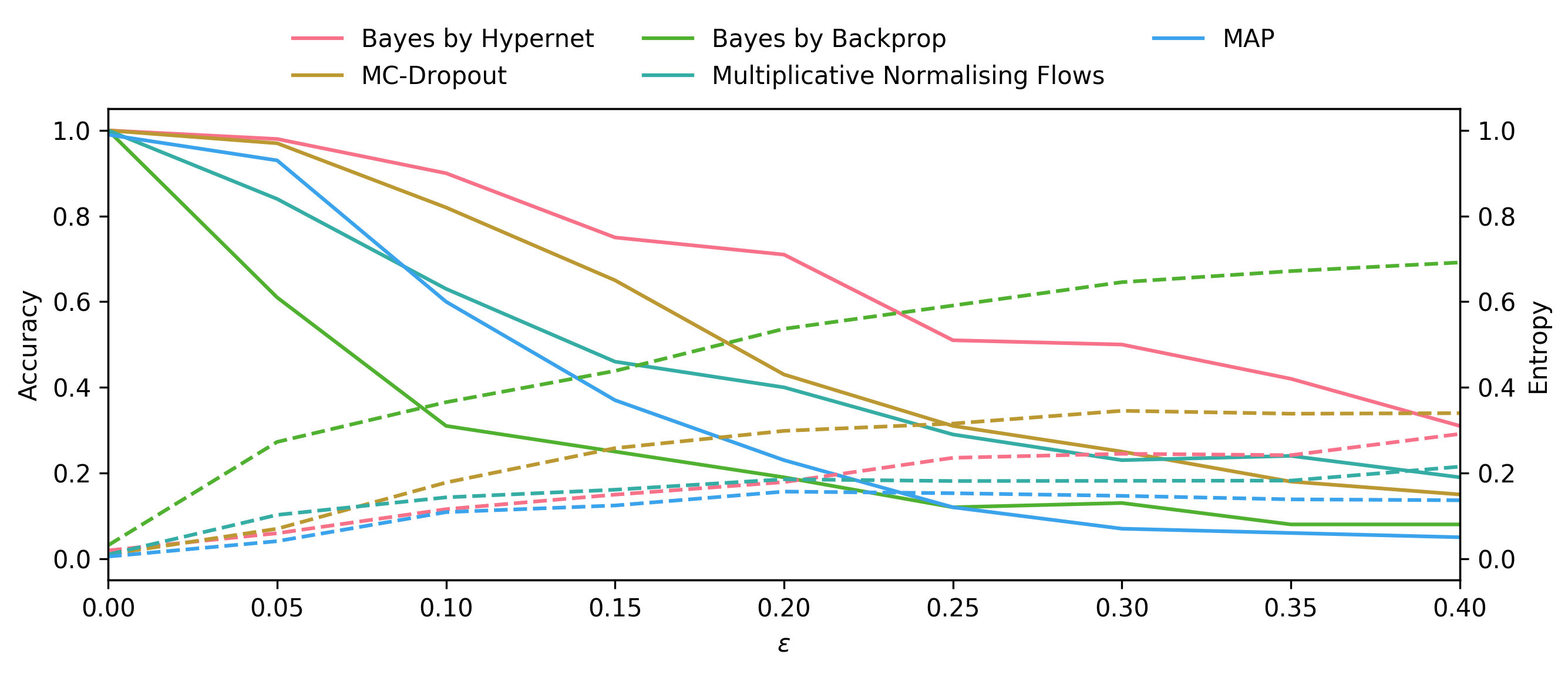}
	\caption{Performance of compared methods exposed to adversarial attacks: The solid line depicts the accuracy and the dashed line the predictive entropy relative to the maximum entropy as a function of the adversarial perturbation.}
    \label{fig:mnist_adv}
\end{figure}

\noindent\textbf{Robustness against adversarial attacks: }
We employ the fast sign method on the first 1000 samples of the MNIST test set. Deep ensembles are excluded, as they are trained on adversarial examples. For variational methods, we generate the adversarial samples as an average of 100 posterior samples to account for the variation during predictions. We plot the accuracy and entropy relative to the maximum entropy in Fig. \ref{fig:mnist_adv}. \emph{BbH} performs the best with the slowest decrease in performance coupled with an increase in predictive uncertainty. The other methods exhibit varying degrees of decay and predictive uncertainty with BbB having the highest uncertainty, but also the steepest decay.

\subsection{Scalability to deep architectures}
To test the scalability of \emph{BbH}, we run experiments using a ResNet-32 \cite{he2016deep} on the same CIFAR5 task as in \cite{louizos2017multiplicative}. It trains on the first five classes of CIFAR10 and uses the remaining as outlier dataset. Apart from MC-Dropout, this is the first reported experiment of Bayesian methods on modern neural network architectures like ResNets. The results in Table~\ref{tab:cifar5} show that deep ensembles achieve the best accuracy. All other methods apart from Dropout exhibit similar accuracy performance and predictive uncertainties. Additionally, we find that the induced overhead of weight generation for both MNF and \emph{BbH} is comparatively small on this larger network. This means that \emph{BbH} is only slower than MAP-training and Dropout

\begin{table}[tb]
\centering
\caption{CIFAR5 classification task: Error [\%] shows the classification error on the test set of the first 5 CIFAR10 classes. The CIFAR5 AUC and Outlier AUC refer to the area under the curve of the CDF of the predictive entropy on the corresponding data set. Compared methods: maximum a posteriori training (MAP), deep ensembles \cite{lakshminarayanan2017simple} (Ensembles), Bayes by Backprop \cite{blundell2015weight} (BbB), MC-Dropout \cite{gal2015bayesian} (Dropout), Multiplicative Normalizing Flows \cite{louizos2017multiplicative} (MNF) and Bayes by Hypernet (\emph{BbH}).\\}
\label{tab:cifar5}
\begin{small}
	\begin{tabular}{ l r r r r}
	\toprule
 &  Error [$\%$]& CIFAR5 AUC & Outlier AUC & Runtime [s]\\
 \midrule
 MAP & $ 13.58 $ & $ 0.83 $ & $ 0.68 $ & $5269$ \\
 Ensemble & $ 10.18 $ & $ 0.77 $ & $ 0.55 $ & $51755$ \\
 BbB & $ 13.78 $ & $ 0.65 $ & $ 0.44 $ & $10562$ \\
 Dropout & $ 24.74 $ & $ 0.46 $ & $ 0.35 $ & $6189$ \\
 MNF & $ 12.82 $ & $ 0.82 $ & $ 0.62 $ & $15743$ \\
 BbH (Ours) & $ 13.62$ & $ 0.68$ & $ 0.51$ & $7896$ \\
  \bottomrule
\end{tabular}
\end{small}
\end{table}

\begin{figure}[tb]
\centering
	\includegraphics[width=\linewidth]{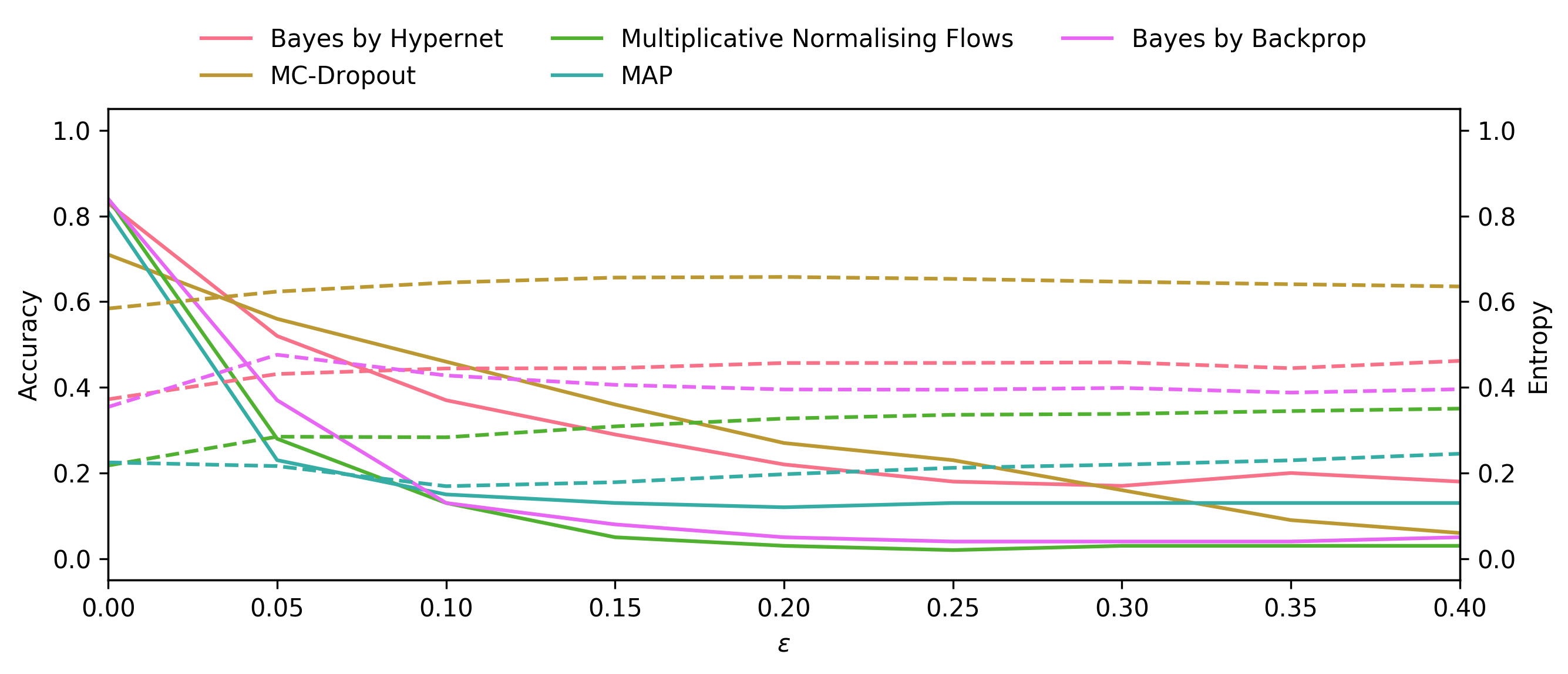}
	\caption{Performance of the methods exposed to adversarial attacks in the CIFAR5 domain: The solid line shows the accuracy and the dashed line the predictive entropy relative to the maximum entropy as a function of the adversarial perturbation.}
    \label{fig:cifar5_adv}
\end{figure}

Performing the same adversarial robustness experiment as with MNIST (see Fig. \ref{fig:cifar5_adv}), we find that \emph{BbH} is again the most robust against adversarial attacks. Only Dropout has a better performance for small $\epsilon$, but performs significantly worse on high $\epsilon$.

\subsection{Examining the posterior distributions}
\begin{figure}[tb]
\centering
\begin{subfigure}[b]{1.\textwidth}
\includegraphics[width=\linewidth]{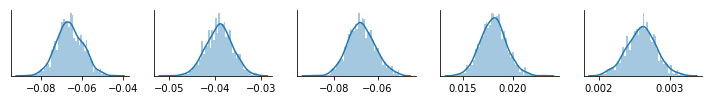}
\caption{Examples of the posterior weight distributions of the LeNet using MNF.}
\end{subfigure}
\\
\begin{subfigure}[b]{1.\textwidth}
\includegraphics[width=\linewidth]{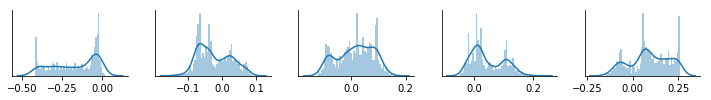}
\caption{Examples of the posterior weight distributions of the LeNet using BbH.}
\end{subfigure}
	\caption{Illustration of the posterior distributions approximated by MNF (a) and BbH (b). BbH clearly generates more complex approximations whereas MNF's resemble Gaussians.}
    \label{fig:weights}
\end{figure}

We examine the fitted posterior distributions of \emph{BbH} and compare them to those of MNF (Fig. \ref{fig:weights}). We find that even though normalising flows are capable of modelling highly complex distributions, the variational posterior still closely resembles a Gaussian. We attribute this to the multiplicative nature of MNF with the underlying Gaussian distributions. \emph{BbH} fits highly complex multi-modal distributions. Furthermore, we examine the correlations\footnote{See Fig. \ref{fig:supp_mnist_corr} and \ref{fig:supp_cifar_corr} in the supplementary material.} of the generated weights of the first convolutional kernel. We find that MNF fails to model correlations between the weights whereas the majority of \emph{BbH} weights are correlated.

\section{Discussion \& Conclusion}
\emph{BbH} interprets \textit{hypernetworks} \cite{ha2016hypernetworks} as an implicit distribution, which we employ as approximate distribution within variational inference. In our experiments, we demonstrate that \emph{BbH} yields strong predictive performances with competitive uncertainties. \emph{BbH} finds a good trade-off between accuracy and predictive uncertainty on MNIST and CIFAR tasks and is the most robust method against adversarial examples. Compared to other Bayesian methods \emph{BbH} yields comparable or better accuracies, while providing superior uncertainties.

Additionally, this work is the first to report an extensive comparison on Bayesian methods for ResNet architectures. It demonstrates the superior scalability of \emph{BbH} to larger network architectures compared to the other Bayesian methods while maintaining competitive accuracy, predictive uncertainty and the best robustness against adversarial attacks with low runtime. \emph{BbH} qualitatively produces more complex approximative posterior distributions (Fig.~\ref{fig:weights}) compared to MNF, which should enable similarly complex distributions. This translates to the complex correlations of the weights that are modelled by \emph{BbH}\footnote{See Sec.~\ref{sec:supp_posterior} for further posterior distributions and the corresponding correlations.}.
Interestingly, we find that our baseline implementations perform better on MNIST than reported in \cite{louizos2017multiplicative}. This suggests that many of those methods require more careful hyperparameter tuning to offer reliable comparisons. Furthermore, even though \emph{BbH} clearly models more complicated posterior distributions than MNF or \emph{BbB}, it does not always yield the highest predictive uncertainties. This raises questions whether richer variational approximations always lead to better results and should be further investigated.

\textbf{Future directions:} We believe, that this work opens a wide variety of directions for future studies: Naturally, the parametrisation of the weights of the main network could be extended to find more efficient and richer forms. This might extend to dynamic \textit{hypernetworks}, which generate weights conditioned on the input to the main network. \emph{BbH} enables the use of highly complex priors as it is merely required to sample from them (\emph{e.g.} task-specific priors instead of Gaussian priors, which are subject to known limitations \cite{neal1995phd}, can be examined). This idea can be extended to transfer learning, where not only the weights, but also previously trained posterior distributions can be transferred and used as a new prior. Additionally, it would be interesting to combine \emph{BbH} with neural architecture search methods like SMASH \cite{brock2017smash} to build Bayesian approximations of the posterior over the architectures.

\textbf{Conclusions: } In this paper, we proposed and extensively evaluated \textit{Bayes by Hypernet}, a new approach to obtaining uncertainty estimates with neural networks. The appropriation of \textit{hypernetworks} to generate weight distributions allows for modelling arbitrary complex distributions and the proposed method naturally integrates with modern deep learning, addressing the need for certainty measures in real-world applications.     

\section*{Acknowledgements}
We like to thank Miguel Jacques, Pierre Richemond, Elliot Crowley, and Joseph Mellor for insightful discussions and comments on the paper. NP is supported by Microsoft Research PhD Scholarship and the EPSRC Centre for Doctoral Training in High Performance Embedded and Distributed Systems (HiPEDS, Grant Reference EP/L016796/1). MR is supported by an Imperial College Research Fellowship. We gratefully acknowledge the support of NVIDIA with the donation of one Titan X GPU for our research. This project has received funding from the European Research Council (ERC) under the European Union's Horizon 2020 research and innovation programme (grant agreement No 757173, project MIRA, ERC-2017-STG).

\bibliography{biblio}
\bibliographystyle{plain}

\appendix

\section{Evaluating Adversarial Variational Bayes and the Kernel KL}\label{sec:supp_bbb}
%\martin{insert a separate bibliography for citations in the supplemental docs, however check that it is covered by the lit in the main paper.}
We run Bayes by Backprop to compare the results of approximating the KL divergence using an adversarial approach or kernel-based approach with the analytical solution. We use the same settings for BbB as in the main paper and train the discriminator for 100 steps before starting training of the main network and then train it for 5 steps for every step we train the main network. We find that the kernel estimation with independent treatment of the weights achieves results closest to the analytical ones. AVB, however, provides the worst accuracy, most overconfident predictions as well as longest runtime.

\begin{table}[H]
\centering
\caption{Comparing AVB and a kernel-based estimations of the KL divergence: AVB produces not only worse accuracies but also the most overconfident results while having the longest runtime. The kernel estimates achieve results close to the analytical with the independent treatment of each weight achieving the closest result to the analytical approach.\\}
\label{tab:bbb_vi}
\begin{small}
	\begin{tabular}{l r r r r}
	\toprule
 &  Error [$\%$]& MNIST AUC & Outlier AUC & Runtime [s]\\
 \midrule
 Analytical & $0.72$ & $0.97$ & $0.41$ & $2892$ \\
 Adversarial Variational Bayes & $0.97$ & $1.00$ & $0.95$ & $22187$\\
 Kernel with independent prior & $0.76$ & $0.98$ & $0.46$ & $5707$ \\
 Kernel with full prior & $0.88$ & $0.99$ & $0.58$ & $4644$\\
  \bottomrule
\end{tabular}
\end{small}
\end{table}

\section{Posterior Distributions}\label{sec:supp_posterior}
\subsection{LeNet on MNIST}
\begin{figure}[hbt]
\centering
\begin{subfigure}[b]{1.\textwidth}
\includegraphics[width=\linewidth]{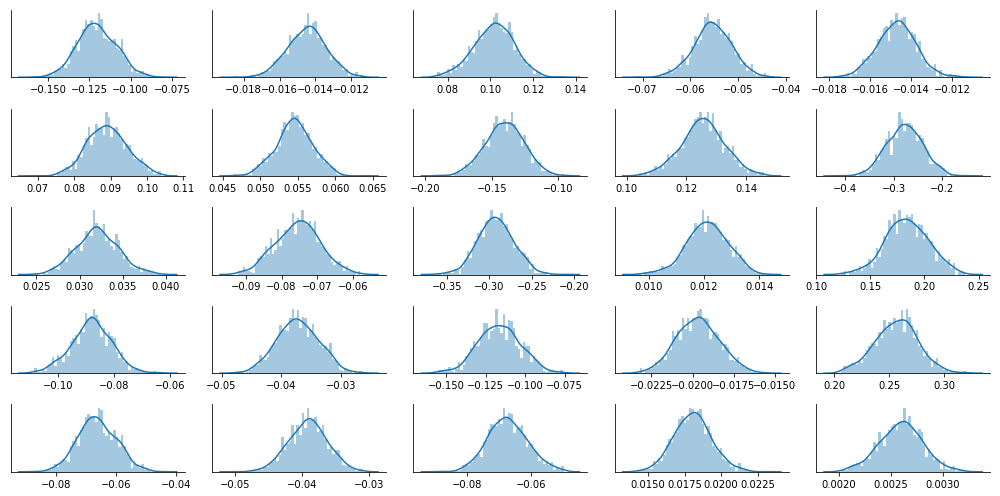}
\caption{Examples of the posterior weight distributions of the LeNet using MNF.\\}
\end{subfigure}
\\
\begin{subfigure}[b]{1.\textwidth}
\includegraphics[width=\linewidth]{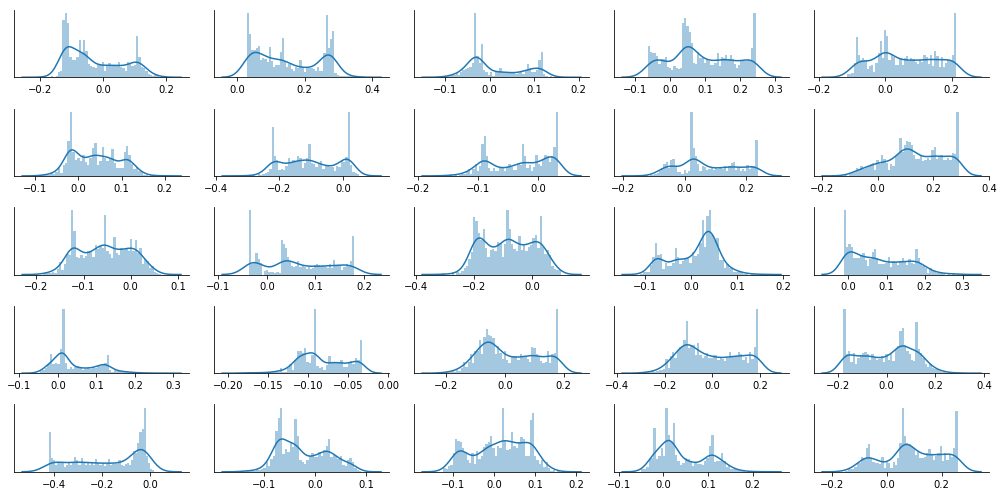}
\caption{Examples of the posterior weight distributions of the LeNet using BbH.}
\end{subfigure}
	\caption{Illustration of the posterior distributions of the 25 first weights of a LeNet trained on the MNIST digit classification task approximated by MNF (a) and BbH (b). BbH clearly generates more complex approximations whereas MNF's resemble Gaussians.}
    \label{fig:supp_mnist_weights}
\end{figure}

\begin{figure}[hbt]
\centering
\begin{subfigure}[t]{0.45\textwidth}
\includegraphics[width=\linewidth]{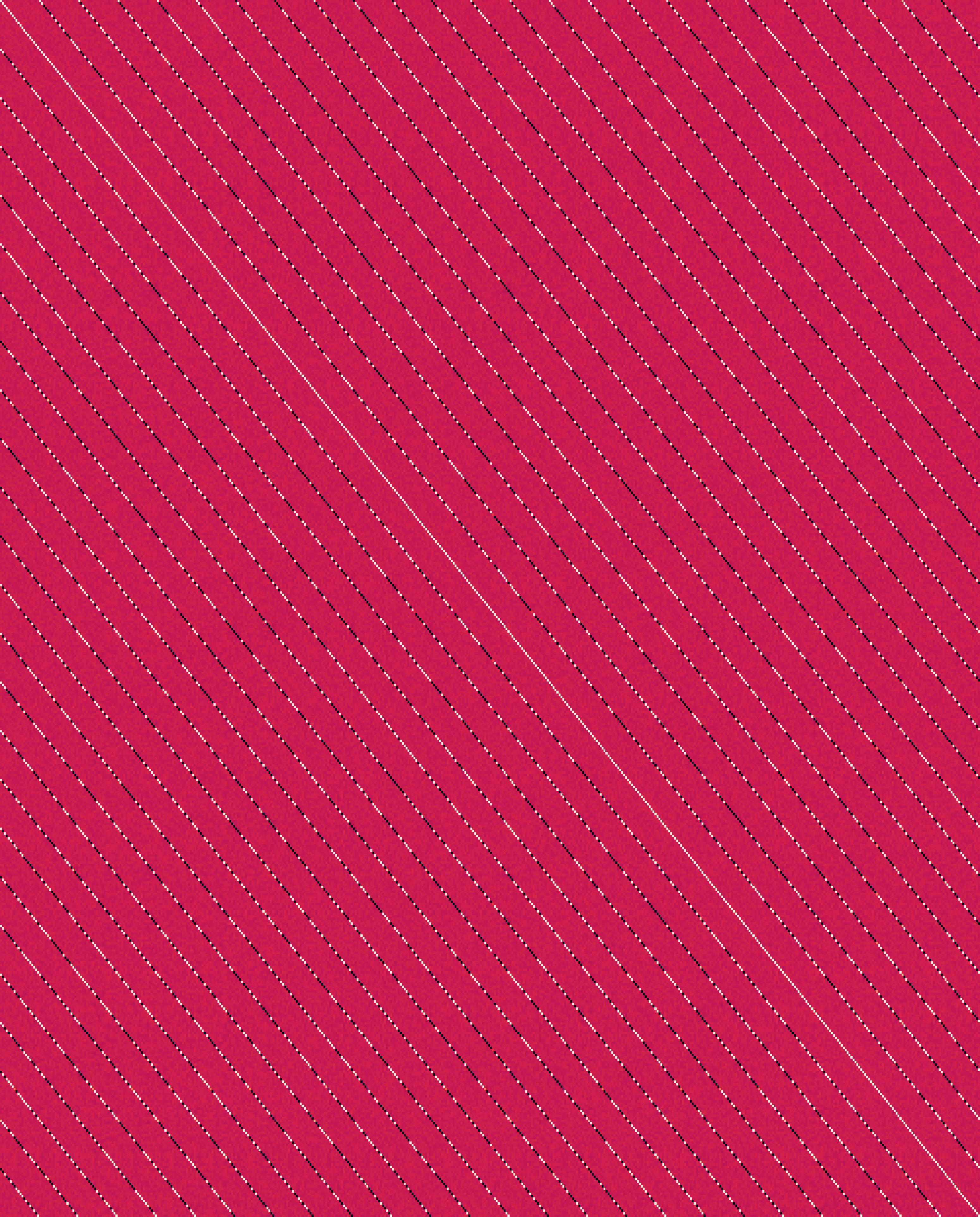}
\caption{Correlations modelled by MNF.}
\end{subfigure}
\hfill
\begin{subfigure}[t]{0.45\textwidth}
\includegraphics[width=\linewidth]{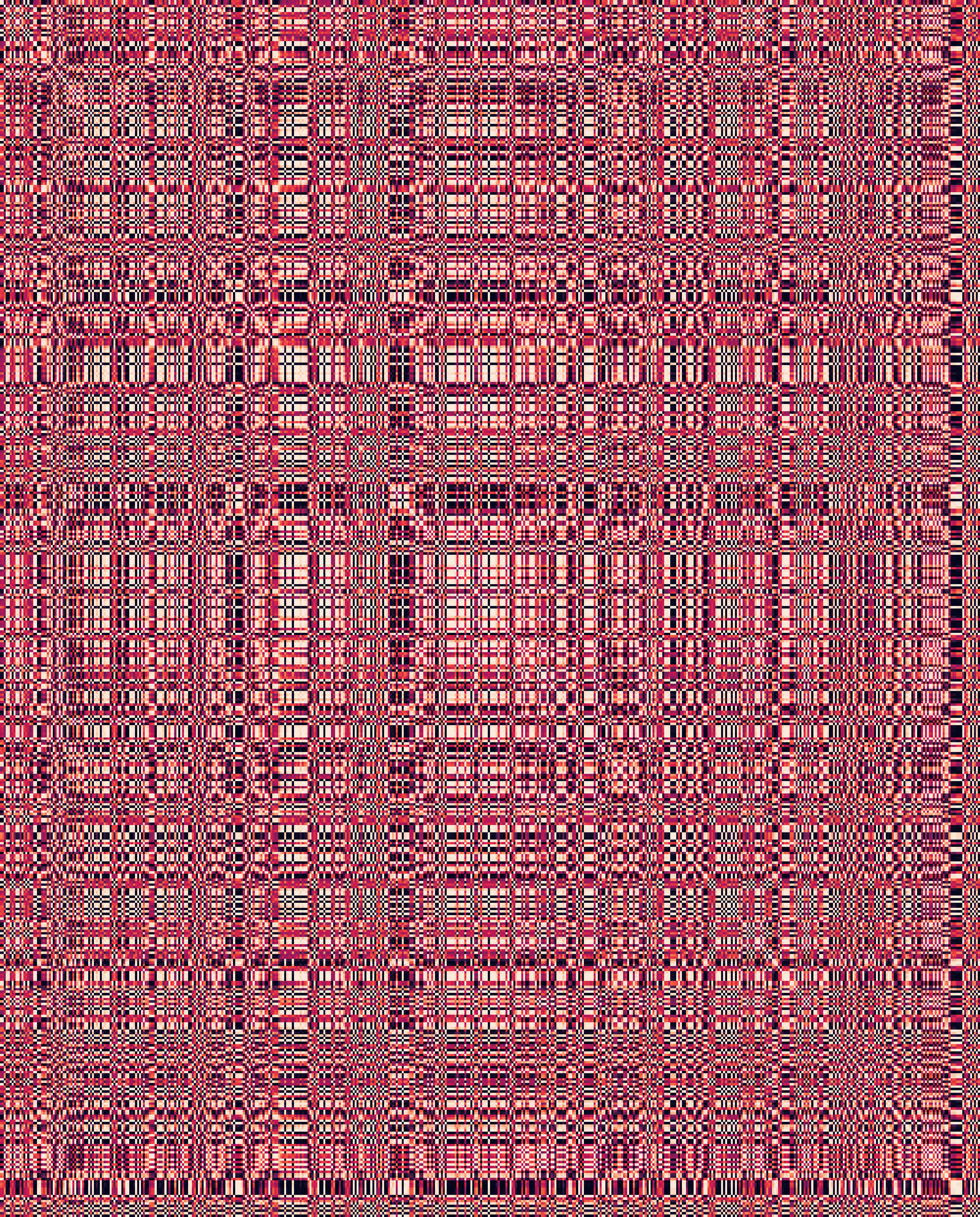}
\caption{Correlations modelled by \emph{BbH}.}
\end{subfigure}
	\caption{Illustration of the correlations between the weights in the first convolutional layer of a LeNet trained on the MNIST digit classification modelled by the posterior distributions approximated by MNF (a) and \emph{BbH} (b). Dark spots indicate negative and bright spots positive correlation. \emph{BbH} models complex dependencies between the weights whereas MNF is only capable of modelling dependencies along the dimension of the multiplicative factor.}
    \label{fig:supp_mnist_corr}
\end{figure}

\subsection{ResNet-32 on CIFAR-5}
\begin{figure}[hbt]
\centering
\begin{subfigure}[b]{1.\textwidth}
\includegraphics[width=\linewidth]{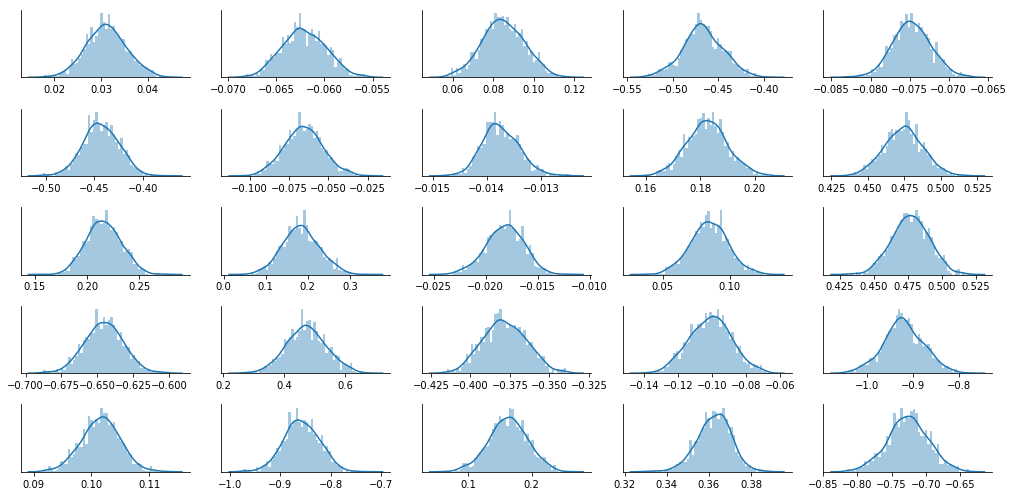}
\caption{Examples of the posterior weight distributions of the LeNet using MNF.\\}
\end{subfigure}
\\
\begin{subfigure}[b]{1.\textwidth}
\includegraphics[width=\linewidth]{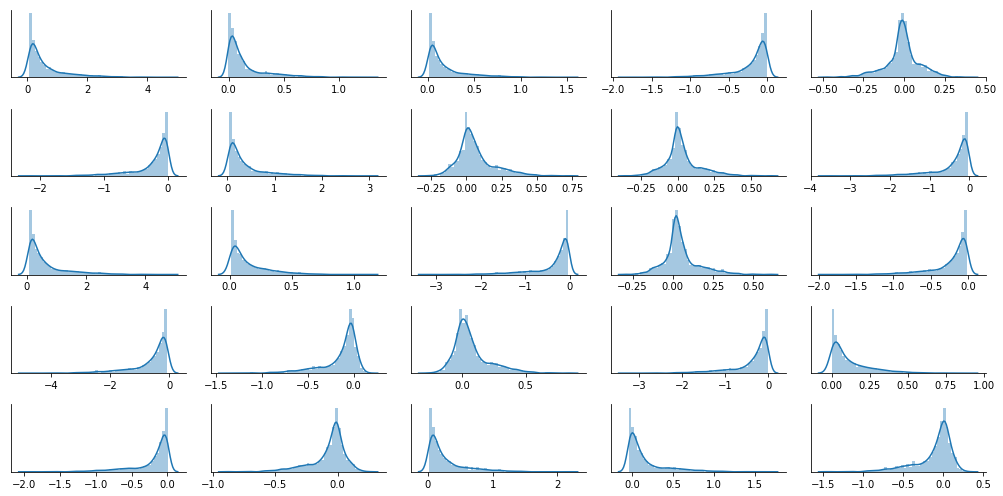}
\caption{Examples of the posterior weight distributions of the LeNet using BbH.}
\end{subfigure}
	\caption{Illustration of the posterior distributions of the 25 first weights of a ResNet-32 trained on the CIFAR-5 classification task approximated by MNF (a) and BbH (b). MNF models posterior distributions that resemble Gaussians. \emph{BbH} models distributions that are more complex than those of MNF but less than the ones it modelled for the MNIST task.}
    \label{fig:supp_cifar_weights}
\end{figure}

\begin{figure}[hbt]
\centering
\begin{subfigure}[t]{0.45\textwidth}
\includegraphics[width=\linewidth]{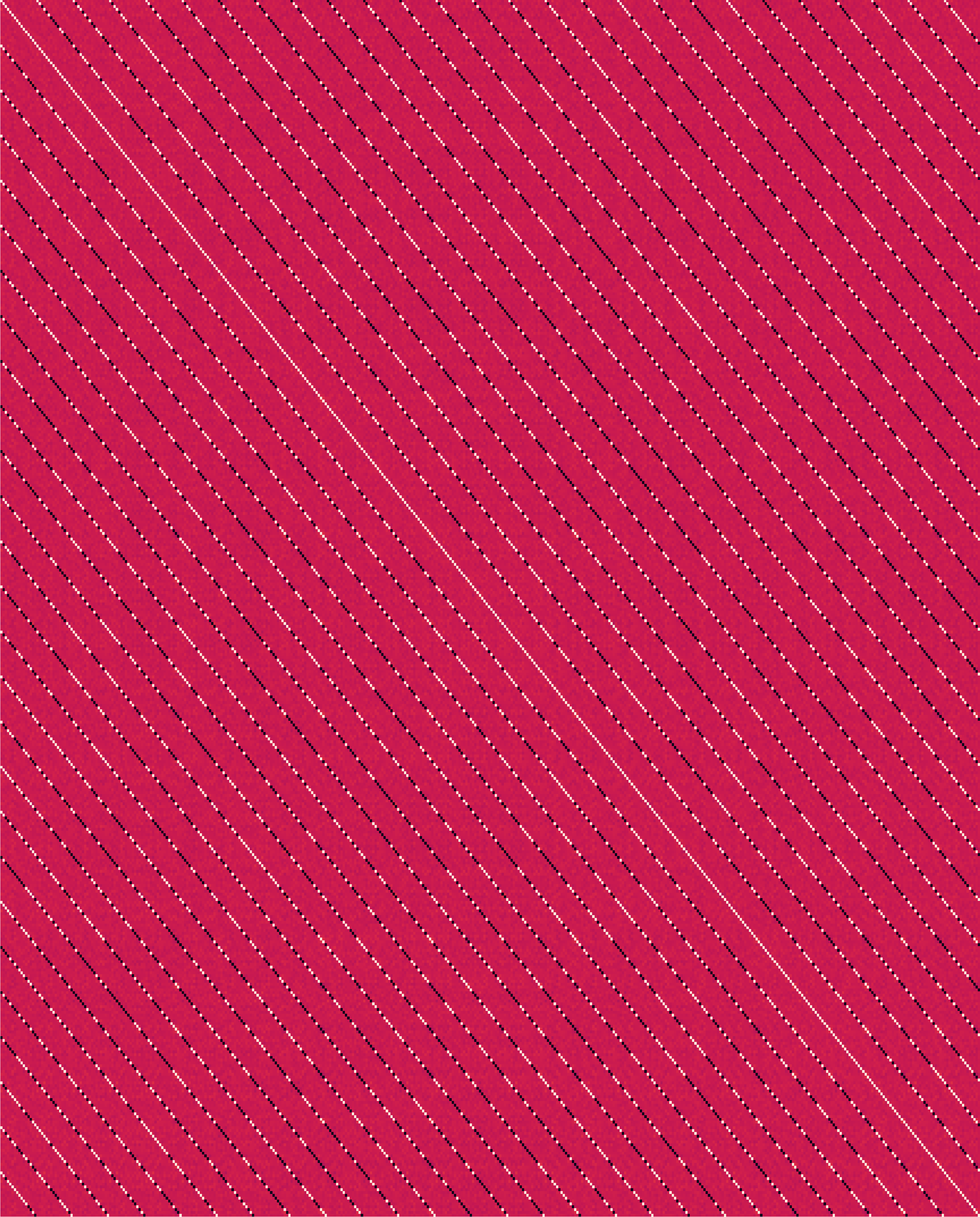}
\caption{Correlations modelled by MNF.}
\end{subfigure}
\hfill
\begin{subfigure}[t]{0.45\textwidth}
\includegraphics[width=\linewidth]{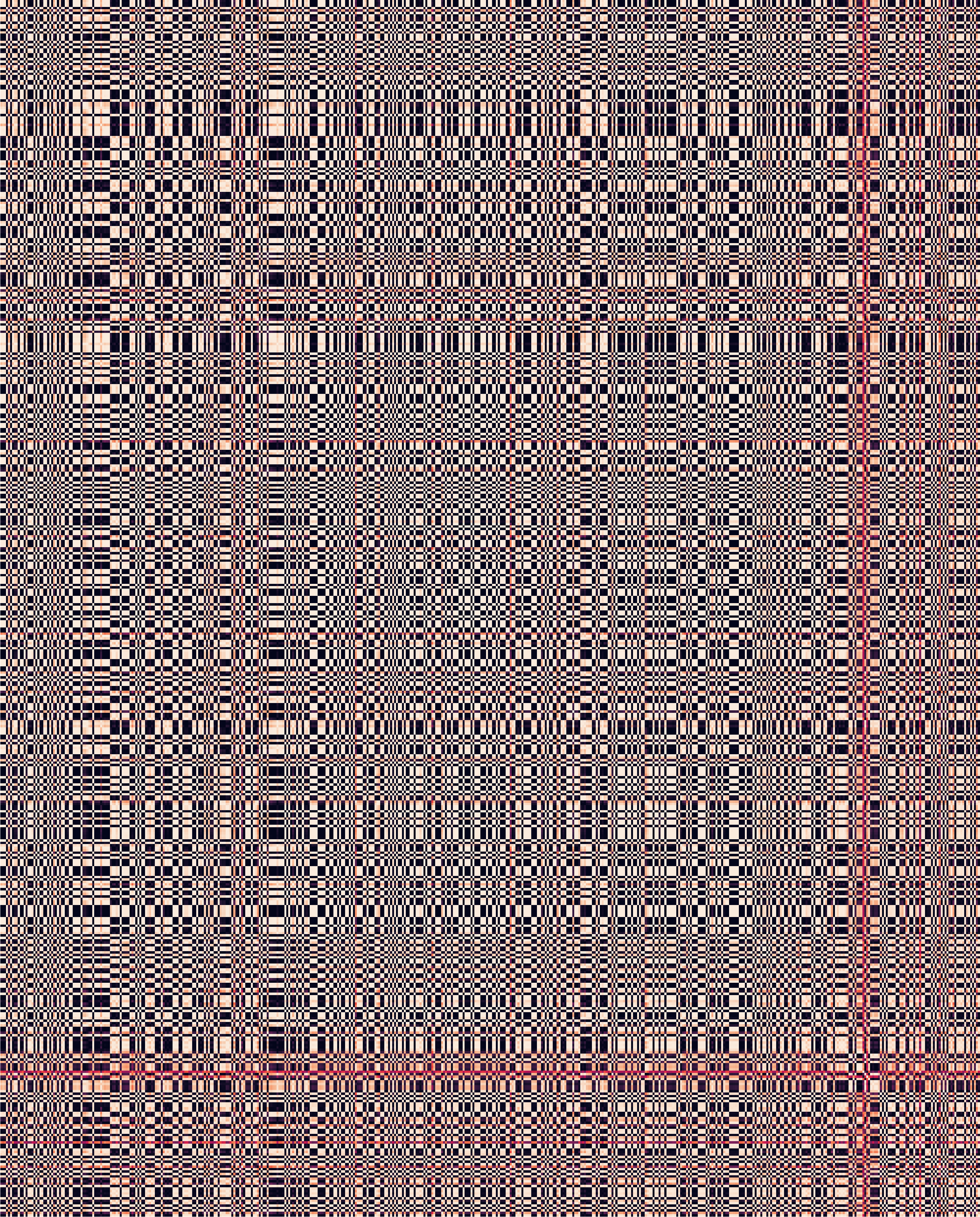}
\caption{Correlations modelled by \emph{BbH}.}
\end{subfigure}
	\caption{Illustration of the correlations between the weights in the first convolutional layer of a ResNet-32 trained on the CIFAR-5 classification task modelled by the posterior distributions approximated by MNF (a) and \emph{BbH} (b). Dark spots indicate negative and bright spots positive correlation. \emph{BbH} models complex dependencies between the weights whereas MNF is only capable of modelling dependencies along the dimension of the multiplicative factor.}
    \label{fig:supp_cifar_corr}
\end{figure}

\end{document}